\documentclass[10pt,twocolumn,letterpaper]{article}

\usepackage{cvpr}              
\definecolor{cvprblue}{rgb}{0.21,0.49,0.74}
\usepackage[pagebackref,breaklinks,colorlinks,allcolors=cvprblue]{hyperref}

\def\paperID{26235} 
\def\confName{CVPR}
\def\confYear{2026}

\title{Contact Matrix: Enhancing Dance Motion Synthesis \\ with Precise Interaction Modeling}

\author{Xuhai Chen$^1$
~ ~ Zhi Cen$^1$
~ ~ Huaijin Pi$^2$
~ ~ Sida Peng$^1$
~ ~ Xiaowei Zhou$^1$
~ ~ Yong Liu$^1$\thanks{Corresponding author.} \\
\normalsize $^1$ Zhejiang University ~ ~ $^2$ The University of Hong Kong \\
{\tt\small 22232044@zju.edu.cn, yongliu@iipc.zju.edu.cn} }

\begin{document}
\maketitle
\begin{abstract}
Generating realistic reactive motions, in which one person reacts to the fixed motions of others, is challenging due to strict interaction constraints and a limited feasible solution space. 
This paper focuses on a typical scenario: duet dance, where high-quality data is scarce, motion patterns are complex, and the details of human interactions are both intricate and abundant. 
To tackle these challenges, we propose a novel two-stage framework. 
In the first stage, we introduce a motion VQ-VAE with separate body-part encoders and a joint decoder, enabling specialized codebooks to enhance representation capacity while dynamically modeling dependencies across body parts during decoding, thereby preventing inconsistencies in the generated motions.
In the second stage, we propose a contact-aware diffusion model for reactive motion generation that jointly generates motion and a contact matrix between individuals, enabling explicit interaction modeling and providing guidance toward more precise and constrained interaction dynamics during sampling.
Experiments show that our method outperforms Duolando with lower 
$\text{FID}_{k}$ (8.89 vs. 25.30) and $\text{FID}_{cd}$ (8.01 vs. 9.97), 
as well as a higher BED (0.4606 vs. 0.2858), indicating improved interaction fidelity and rhythmic synchronization.
\end{abstract}    
\section{Introduction}
\label{sec:intro}

Motion synthesis has been widely explored across diverse domains, including animation, AR/VR, and robotics~\cite{hanser2010scenemaker, egges2007presence, yoon2019robots, guo2022generating, kim2022brand, alexanderson2023listen}.
Although the field has seen significant progress, most existing studies focus on generating motions for a single character conditioned on external inputs.
These inputs are typically static and non-interactive, such as textual descriptions, audio signals, or scene contexts~\cite{yu2020structure, t2mgpt, le2023music, wu2022saga}.
In contrast, real-world human motion often takes place in multi-person scenarios~\cite{duolando, regennet, remos, synergy}, where individuals continuously adapt their movements in response to one another.
This dynamic and interdependent process is commonly referred to as reactive motion.

Reactive motion synthesis faces two major challenges.
First, data for multi-person interactions is harder to collect than for individual actions and often involves more subtle and complex movements.
Second, physical contact occurs more frequently and with greater intricacy in interactive settings, where accurate body contact is crucial for conveying intent, and even minor errors can easily disrupt realism.
To study reactive motion generation, we focus on duet dance, a representative scenario in which a follower reacts to a leader.
In this setting, interactions are frequent and in close proximity, further amplifying the two challenges.
\begin{figure}[t]
  \centering
  \includegraphics[width=.92\linewidth]{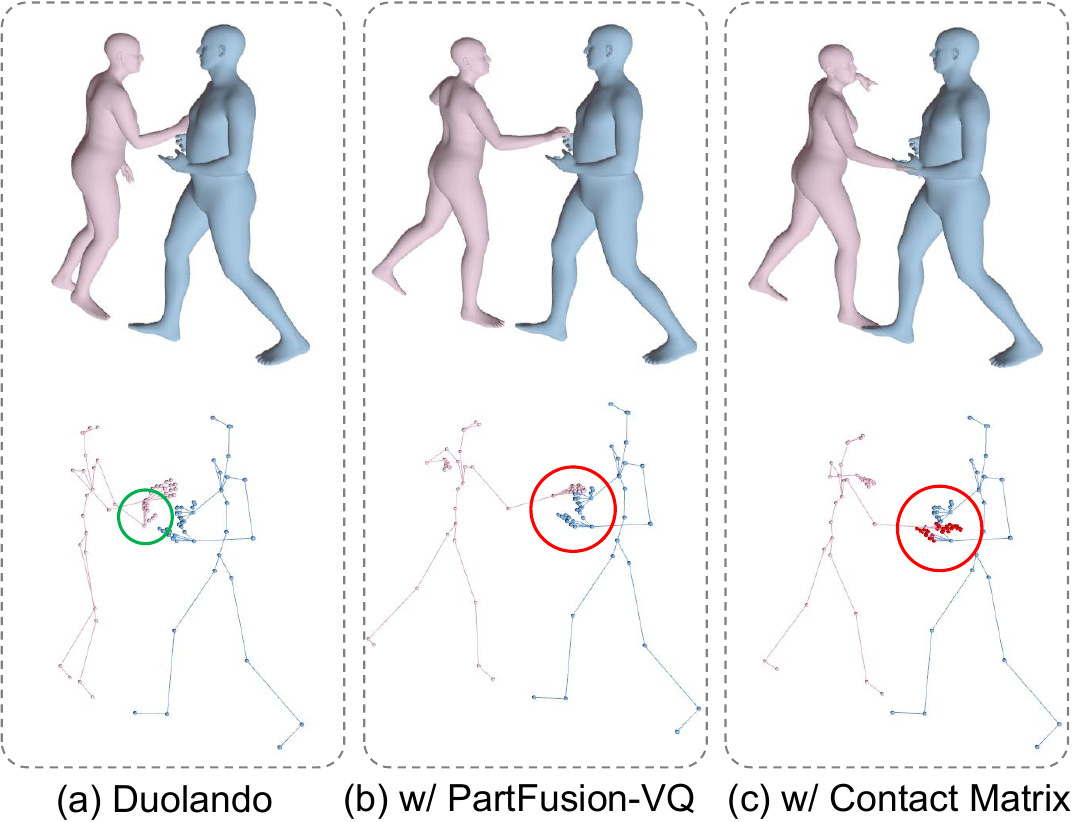} 
  \caption{\textbf{We tackle interaction-aware reactive motion generation.} First, we introduce (b) PartFusion-VQ to prevent unnatural poses, as highlighted by the green circle in (a) Duolando~\cite{duolando}. Second, in (c), we incorporate a contact matrix to guide motion refinement, correcting unrealistic joint configurations (red circles).}
  \label{fig:intro}
  \vspace{-1em}
\end{figure}

For the first challenge, we aim to make more effective use of limited data through a compact and structured motion representation.
Recently, many methods use VQ-VAE~\cite{vqvae} to encode motion into a discrete latent space~\cite{t2mgpt, motiongpt, synergy}, enabling generation from latent codes instead of raw joint representations~\cite{interformer, raig, li2017auto}.
This latent representation implicitly captures key physical constraints such as joint angle limits, bone-length consistency, and gravity~\cite{akhter2015pose, pawar2008content, pradhan2007hierarchical}, resulting in more plausible and stable motions~\cite{li2017auto}.
However, in reactive motion generation, where data is limited and movements are more complex, learning a well-structured latent space becomes particularly challenging.
To better capture part-specific motion patterns under such conditions, some studies~\cite{bailando, duolando} divide the body into multiple parts and train separate VQ-VAEs for each.
While this design allows for finer control and representation of motion, it often results in incoherent global dynamics when the parts are combined, as shown by the green circle in~\cref{fig:diff} (a)..
In light of this limitation, we propose \textbf{PartFusion-VQ}, which retains a dedicated codebook for each body part to preserve local expressiveness, while fusing all part-wise features through a shared decoder to ensure global coherence and diversity under limited data.

To address the second challenge and improve interaction realism, we incorporate physical contact as a guiding factor.
Most existing reactive motion generation methods, in both two-person~\cite{duolando, remos, regennet, interformer} and multi-person~\cite{jeong2024multi} settings, ignore contact by predicting local motion and global trajectory separately and simply combining them.
This often results in unrealistic outcomes, such as a large distance between characters performing close-contact actions and physical collisions despite the absence of any interaction cues.
Motivated by this and drawing inspiration from the method in~\cite{ma2024contact}, we extend the framework to additionally generate a joint-level contact matrix alongside local motion and global trajectory.
Specifically, in our duet dance setting, we adopt a conditional diffusion framework~\cite{sohl2015deep, ddpm, ddim} to jointly generate local motion and global trajectory for the follower, along with a joint-level contact matrix representing interactions between follower and leader, conditioned on leader motion and background music.
To further ensure physically plausible interactions, we incorporate a contact-guided refinement mechanism during diffusion inference~\cite{yang2024smgdiff, song2020score}. 
When contact is predicted between specific joints, the guidance function adjusts the sampling updates of the diffusion model, steering the generated motions toward satisfying these contact constraints, as indicated by the red circles in~\cref{fig:intro} (b) and (c).
This process encourages both realistic local motion and coherent global spatial coordination. We refer to this diffusion-based \textbf{R}eactive \textbf{C}ontact-aware motion generation model as \textbf{RCDiff}.


Our contributions can be summarized as follows:
\begin{itemize}
    \item To address the challenge of learning diverse motion with limited data, we propose PartFusion-VQ, which encodes body parts separately and decodes them jointly, capturing both part-specific diversity and full-body coherence.
    \item To improve interaction realism, we introduce RCDiff, a conditional diffusion framework that generates follower local motion, global trajectory, and a joint-level contact matrix, with a contact-guided refinement strategy during inference to enforce physically plausible interactions.
    \item Extensive experiments show that our method surpasses existing approaches. Compared to Duolando, it achieves lower $\text{FID}_{k}$ (8.89 vs. 25.30), $\text{FID}_{cd}$ (8.01 vs. 9.97), and higher BED (0.4606 vs. 0.2858), highlighting improvements in coordination and rhythm.
\end{itemize}

\section{Related Works}

\subsection{Reactive Motion Generation}

Reactive motion generation focuses on synthesizing the motion of one individual conditioned on the fixed motions of others. This setting differs from joint multi-person motion generation~\cite{ghosh2025duetgen, liang2024intergen}, where all motions are generated simultaneously and co-adapted for global consistency.

Most prior work on reactive motion generation focuses on two-person interactions across various scenarios~\cite{duolando, regennet, remos, synergy, readytoreact}. Early methods, such as Mousas et al.~\cite{mousas2018performance}, formulate reactive motion generation as a decoding problem using a Hidden Markov Model (HMM) to control a digital partner in duet dance, while DRAM~\cite{ahuja2019react} predicts a primary motion sequence for the speaker and refines it based on the posture and speech of the interlocutor. In basketball scenarios, Starke et al.~\cite{starke2020local} employ a Mixture of Experts model with local phase functions and joint distance constraints to improve interactive realism.

More recent Transformer-based approaches~\cite{transformer} further advance interactive motion modeling by capturing temporal and spatial dependencies across joints and agents. InterFormer~\cite{interformer} uses attention mechanisms on skeletal graphs and interaction distances to enhance cross-agent coordination. ReMos~\cite{remos} and ReGenNet~\cite{regennet} integrate diffusion-based decoders to enable hierarchical or real-time reactive generation, while Duolando~\cite{duolando} and Maluleke et al.~\cite{synergy} encode motion with VQ-VAE and autoregressively generate follower sequences with Transformers.
Despite these advances, existing methods often struggle to capture fine-grained interactive cues, which are especially crucial in scenarios involving close human-to-human interactions.

Therefore, to explicitly model and leverage such interaction details, we introduce contact matrices in the two-person reactive motion generation task. By predicting and using these contact matrices to guide the generated motions, our method improves physical plausibility, enhances coordination between participants, and ensures that joints intended to interact are properly aligned.

\subsection{Contact-aware Motion Generation}

Contact information has long been regarded as a crucial factor in motion generation, particularly in human-object interaction (HOI) scenarios~\cite{ma2024contact, diller2024cg, cha2024text2hoi, zhang2022couch}, where it plays a vital role in enhancing physical plausibility and interaction realism.
For instance, CG-HOI~\cite{diller2024cg} proposes a joint diffusion framework that simultaneously models human motion, object motion, and contact distances between body surfaces and objects to improve the coherence of generated interactions. 
Text2HOI~\cite{cha2024text2hoi} introduces a two-stage framework that first predicts hand-object contact maps and then generates motions conditioned on these contacts, demonstrating that contact priors can significantly enhance motion plausibility. 
Similarly, CATMO~\cite{ma2024contact} presents a contact-aware text-to-motion framework that jointly models body motion and contact sequences through dual VQ-VAE and GPT modules.
On the human-scene side, COUCH~\cite{zhang2022couch} leverages contact positions with objects to achieve controllable behaviors like sitting or leaning.

While contact modeling has been widely studied in HOI scenarios, it remains relatively underexplored in reactive motion generation. Unlike HOI, this setting poses new challenges: contacts are more frequent, fine-grained, and diverse, significantly narrowing the feasible interaction space. Moreover, as both individuals actively move, contacts become time-dependent and coupled with motion dynamics such as relative velocity. To address these challenges, we introduce a joint-level contact matrix to represent and guide fine-grained human–human interactions during generation.

\subsection{Motion Latent Representation}
A core problem in human motion research is how motion is represented and parameterized.
Such representations are widely explored across diverse motion-related tasks, including recognition~\cite{yan2018spatial, zhang2020semantics, liu2020disentangling}, retargeting~\cite{aberman2020skeleton, villegas2018neural, aberman2019learning}, prediction~\cite{mao2019learning, mao2020history, dabral2023mofusion}, generation~\cite{t2mgpt, motiongpt, bailando, duolando}, and beyond.

Early approaches typically represent motion in continuous 3D space by directly predicting joint positions or rotations~\cite{6drot}, sometimes parameterized using articulated human body models such as SMPL~\cite{smpl} or SMPL-X~\cite{smplx}.
However, these methods do not explicitly model motion structure or consider constraints such as joint angle limits, bone-length consistency, and gravity~\cite{akhter2015pose, pawar2008content, pradhan2007hierarchical}, which can adversely affect tasks that require motion synthesis, particularly over long sequences~\cite{li2017auto}.

To address these limitations, recent studies explore modeling human motion using latent representations to capture structured and physically plausible patterns~\cite{chen2023executing, barquero2023belfusion, t2mgpt, motiongpt, bailando, duolando}.
For example, MLD~\cite{chen2023executing} applies a VAE to encode motion before performing diffusion, effectively reducing noise and computational cost.
Similarly, BeLFusion~\cite{barquero2023belfusion} leverages a VAE-based representation for stochastic motion prediction, disentangling behavior from pose and motion to generate diverse yet realistic sequences.
T2M-GPT~\cite{t2mgpt} and MotionGPT~\cite{motiongpt} employ VQ-VAE tokenization for autoregressive motion generation with Transformers, capturing complex temporal dependencies.
Moreover, beyond autoregressive models, VQ-VAE representations have also been combined with diffusion models~\cite{readytoreact}, improving generative quality and sample diversity.
When motion data is scarce or highly complex, some methods~\cite{bailando, duolando} decompose motion into body parts, modeling each part with a separate VQ-VAE to enhance latent representations.  
However, they often struggle to maintain overall consistency across the full body.

In this paper, we propose PartFusion-VQ, using separate codebooks for each body part and a shared decoder to reconstruct the full body, allowing fine-grained part-level control while preserving overall coherence.

\section{Method}
\label{sec:formatting}

In this section, we first formalize the task of reactive motion generation in duet dance (Section~\ref{sec:problem}) and then present a two-stage framework.
The first stage learns discrete latent representations through VQ-VAEs, providing a compact basis for subsequent conditional generation (Section~\ref{sec:vqvae}).
The second stage adopts a contact-aware diffusion model that generates follower motion under the conditions of leader motion and music, while applying contact-guided refinement during inference to ensure realistic and coordinated interactions (Section~\ref{sec:diff}).

\subsection{Problem Formulation} \label{sec:problem}

This paper studies leader-conditioned reactive motion generation in duet dance, where the task is to generate follower motion given leader motion and music.

We represent human motion using SMPL-X~\cite{smplx} with \( J = 54 \).
Rather than generating motion in global space, we decompose it into two components: local motion and global trajectory. 
The local motion is denoted as \( p \in \mathbb{R}^{T \times J \times 3} \), where \( T \) is the number of time steps. Each frame contains the 3D joint positions in the local coordinate centered at the root joint.  
For the follower, the global trajectory is represented by relative positions $d \in \mathbb{R}^{T \times 3}$, defined as the offset between the root positions of the leader and the follower at each time step, allowing the follower's global position to be determined from the leader's.
For the leader, the global trajectory is given but not required for modeling, as the follower trajectory is defined relative to the leader.

To better model fine-grained physical interactions between the two dancers, in addition to predicting local motion and global trajectory of the follower, we generate contact matrices \( c \in \{0,1\}^{T \times J' \times J'} \), where each element \( c_{t,ij} \) indicates whether the \( i \)-th joint of the follower is in contact with the \( j \)-th joint of the leader at time step \( t \). A value of 1 represents contact, and 0 indicates no contact.  
Moreover, to reduce noise from minor hand movements and highlight meaningful interactions, we simplify the hand representation when constructing the contact matrix by averaging all finger joints into a single hand joint on each side.  
As a result of this simplification, \( J' = 23 \) in our experiments. 
This explicit modeling of contact can facilitate capturing precise interaction patterns between the two bodies.

The music is represented as \( a \in \mathbb{R}^{T \times C_m} \), where \( C_m = 54 \). These features are extracted from the audio signal using Librosa~\cite{mcfee2015librosa, duolando}.  
Based on the above definitions, the task can be formulated as:
\begin{equation}
(p^f, d, c) = F(p^l, a),
\end{equation}
where $F$ denotes the conditional generation function, and the superscripts $f$ and $l$ denote the follower and leader.

\subsection{PartFusionVQ: VQ-VAE with Part Encoding and Fusion Decoding} \label{sec:vqvae}
\noindent
\textbf{Local motion.} 
Recent motion generation methods often use a VQ-VAE to encode motion into a latent space, representing motion as latent codes during generation~\cite{duolando, synergy, t2mgpt, motiongpt, bailando}.
By learning from data, this latent space captures both physical constraints and domain-specific patterns, enabling the generation of physically plausible motion that reflects the style of the training data.
However, when the dataset is limited and the motion patterns are complex, the learned latent space often fails to represent motion accurately. To address this, some works~\cite{bailando, duolando} divide the human body into separate parts and independently train a VQ-VAE for each part, but this usually results in incoherent overall motion due to the lack of constraints between parts.

\begin{figure}[t]
  \centering
  \includegraphics[width=\linewidth]{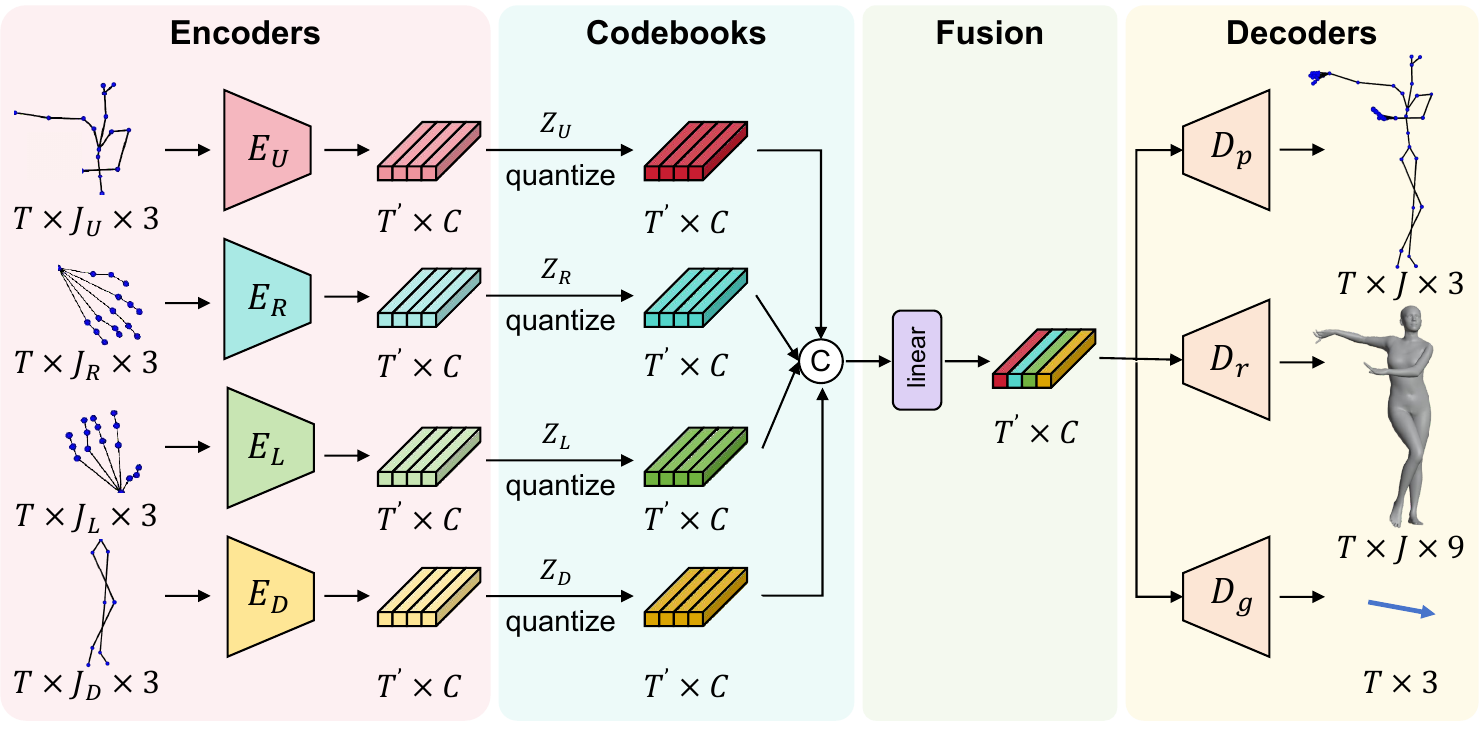} 
  \caption{\textbf{Architecture of PartFusion-VQ.} $J_s$, $E_s$, and $Z_s$ ($s \in \{U, D, L, R\}$) represent the joint number, encoder, and codebook of each body part (upper body, lower body, left hand, right hand), respectively. $T$ and $T'$ denote the input and encoded sequence lengths. $C$ is the feature dimension, and $D_p$, $D_r$, $D_g$ denote the decoders for 3D joint positions, rotations, and global translation.}
  \label{fig:vqvae}
  \vspace{-1em}
\end{figure}

Based on the above observations and inspired by~\cite{pi2023hierarchical}, we propose a new VQ-VAE~\cite{vqvae} structure named PartFusion-VQ, which keeps separate codebooks for each part while using a shared decoder to fuse features and reconstruct the full body. This design preserves the benefits of part-based modeling while enhancing motion consistency. 
Specifically, the local motion \( p \in \mathbb{R}^{T \times J \times 3} \), representing joint positions in a local coordinate system, is divided into four parts: upper body \( p_U \), lower body \( p_D \), left hand \( p_L \), and right hand \( p_R \), as shown in~\cref{fig:vqvae}. 
Each part \( p_s \) with \( s \in \{U, D, L, R\} \) is encoded by a 1D CNN into a deep feature \( z_e^s \in \mathbb{R}^{T' \times C} \), where \( T' = T / f \) is the temporal length after downsampling by a factor \( f \), and \( C \) is the feature dimension.
These features are then quantized by assigning each vector to the nearest entry in the corresponding learned codebook:
\begin{equation}
  \hat{z}_e^s = \arg\min_{z_q^s \in \mathcal{Z}_s} \| z_e^s - z_q^s \|, \quad \forall\, s \in \{U, D, L, R\}.
\end{equation}
Here, \( \mathcal{Z}_s \) denotes the learned codebook for part \( s \), and each \( z_q^s \in \mathcal{Z}_s \) represents a codebook vector.

After quantization, the discrete features \( \hat{z}_e^s \) from the four body parts are fused via a linear layer into a unified representation, which is then decoded into the full-body motion sequence by the decoders.  
In particular, we reconstruct 3D joint positions \( p \) and additionally predict joint rotations \( r \) and global translation \( g \) to enable more accurate motion reconstruction and support mesh-driven visualization~\cite{duolando}. 
As illustrated in~\cref{fig:vqvae}, these outputs are produced by three 1D CNN decoders: \( D_p \), \( D_r \), and \( D_g \), respectively.

We train the model using a composite loss with three parts: reconstruction losses, vector quantization losses, and commitment losses.  
The reconstruction losses minimize \( \ell_1 \)-errors and incorporate first- and second-order derivatives to ensure smooth motion~\cite{bailando, duolando}.  
The vector quantization losses encourage effective usage of the part-specific codebooks, while the commitment losses ensure consistency between the encoder outputs and their assigned codebook entries.  
The overall loss can be formulated as:
\begin{align}  
  \mathcal{L}_{m} &= \mathcal{L}_{\text{rec}} (\hat{p}, p) + \mathcal{L}_{\text{rec}} (\hat{r}, r) + \mathcal{L}_{\text{rec}} (\hat{g}, g) \notag \\  
  &\quad + \sum_{s} \left( \| \text{sg}(z_e^s) - z_q^s \| + \lambda^m \| z_e^s - \text{sg}(z_q^s) \| \right)
\end{align}  
The stop-gradient operation \(\text{sg}(\cdot)\) blocks gradients to allow separate updates of the encoders and the codebooks, and \( \lambda^m \) controls the commitment strength.
\begin{figure*}[t]
  \centering
  \includegraphics[width=0.7\linewidth]{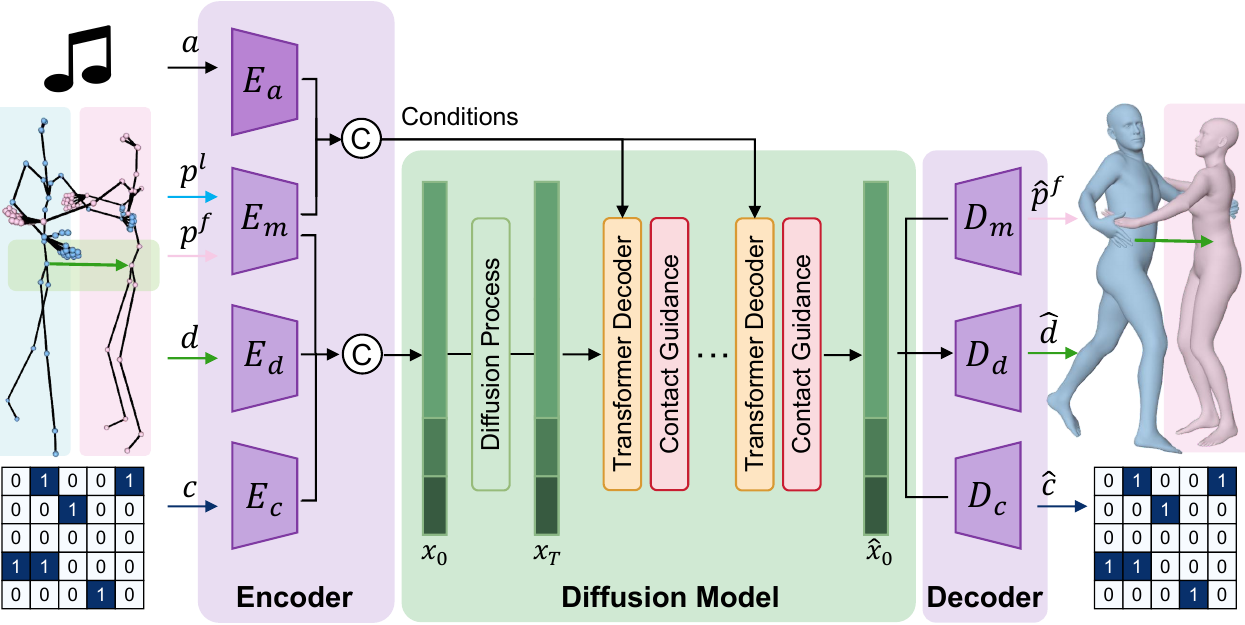} 
  \caption{\textbf{Overview of the pipeline.}  We propose a diffusion model named RCDiff to generate reactive motion in latent space. $E_m$ denotes the four PartFusion-VQ encoders; $E_d$, $E_c$, and $E_a$ denote the encoders for global trajectory, contact matrix, and music, respectively. $D_m$, $D_d$, and $D_c$ denote the corresponding decoders. All encoders and decoders except $E_a$ are frozen during diffusion training.}
  \label{fig:diff}
  \vspace{-1em}
\end{figure*}

\noindent
\textbf{Global trajectory and contact.} 
In addition to local motion, we also encode the follower's global trajectory, represented as relative positions with respect to the leader, and the contact matrix using standard VQ-VAEs.  
This allows all components to be encoded into latent representations, reducing redundancy in low-level inputs and enabling the subsequent generative model to process them in a consistent format.  
We next describe the training objectives for these two VQ-VAEs in detail.  
For clarity, we denote the encoder output, quantized vector, and commitment weight as \( z_e \), \( z_q \), and \( \lambda \), with task-specific superscripts added when needed.

For relative positions \( d \), the VQ-VAE is designed to capture typical patterns of root movement between the leader and the follower.  
Its training objective, similar to that of PartFusion-VQ, combines an \(\ell_1\) reconstruction loss with first- and second-order temporal derivatives, along with vector quantization and commitment losses. 
Formally, the resulting loss can be written as:
\begin{align}
  \mathcal{L}_d &= \mathcal{L}_{\text{rec}}(\hat{d}, d) \notag \\
  &\quad + \| \text{sg}(z_e^d) - z_q^d \| + \lambda^d \| z_e^d - \text{sg}(z_q^d) \|.
\end{align}

For the contact matrix \(c\), the VQ-VAE learns compact embeddings that reflect common joint-level interaction patterns across time. 
To supervise it, we use a focal loss~\cite{lin2017focal}, which focuses on harder examples to address the class imbalance caused by sparse contact events and improve training stability.
In addition, vector quantization and commitment losses are also applied. 
The total loss is defined as:
\begin{align}
  \mathcal{L}_{c} &= \mathcal{L}_{\text{focal}}(\hat{c}, c) \notag \\
                  &\quad + \| \text{sg}(z_e^c) - z_q^c \| + \lambda^c \| z_e^c - \text{sg}(z_q^c) \|.
\end{align}

Notably, as discussed earlier, the local motion VQ-VAE is capable of providing the root translation of the follower, which can be integrated over time from an initial position to reconstruct the global trajectory.  
Thus, explicitly predicting the relative displacement between the follower and the leader for global positioning may seem unnecessary.  
However, integrating predicted translations over time can easily lead to drift due to error accumulation, especially in long sequences.  
In contrast, directly predicting the displacement enables the position of the follower to be computed from the global position of the leader at each time step, thus avoiding error accumulation and improving stability.

\subsection{RCDiff: Diffusion-based Reactive Motion Synthesis with Contact Matrix} \label{sec:diff}

\noindent
\textbf{Conditional Diffusion.} 
Given the latent representations of local motion, global trajectory, and contact matrix obtained from the VQ-VAE encoders described in Section~\ref{sec:vqvae}, we aim to synthesize the follower's motion conditioned on the leader's motion and music features.
To achieve this, we adopt a diffusion-based generative model~\cite{ddpm, regennet, remos} due to its strong capacity for modeling complex temporal dynamics.
Formally, the task is defined as conditional generation in the latent space, modeled by maximizing the likelihood:
\begin{align}
(z_f^\ast, z_d^\ast, z_c^\ast) = \arg\max_{(z_f, z_d, z_c)} P(z_f, z_d, z_c \mid z_l, f_a),
\end{align}
where we use the symbol \(z\) to denote features extracted by VQ-VAE encoders before codebook quantization. 
In this way, the results generated by the diffusion model can be subsequently quantized and decoded to reconstruct follower motion sequences.
Specifically, \(z_f = (z_f^U, z_f^D, z_f^L, z_f^R)\) and \(z_l = (z_l^U, z_l^D, z_l^L, z_l^R)\) denote the follower and leader motion features, including the upper body, lower body, left hand, and right hand. \(z_d\) represents the latent feature of relative positions between the two dancers, and \(z_c\) encodes contact information. 
The music feature \(f_a\) is extracted from the input audio \(a\) using a music encoder \(E_a\). The encoder first embeds the audio features through a linear projection and positional encoding, and then models temporal dependencies using a stack of Transformer blocks~\cite{transformer, duolando}.
Finally, to match the temporal resolution of the VQ-VAE features, a downsampling layer reduces the length of the music sequence accordingly.
All features extracted by the VQ-VAE and music encoders share the same shape of \(T' \times C\).

An overview of the entire generative diffusion pipeline is illustrated in~\cref{fig:diff}.
Accordingly, we define the input sequence \(x = \text{concat}(z_f, z_d, z_c)\) and the condition sequence \(y = \text{concat}(z_l, f_a)\), where all components are aligned and concatenated along the temporal dimension. 
During training, Gaussian noise is progressively added to \(x\) through a Markovian forward process defined as:
\begin{equation}
q(x_t \mid x_{t-1}) = \mathcal{N}(x_t; \sqrt{\alpha_t} x_{t-1}, (1 - \alpha_t) I)
\end{equation}
where \(t \in \{1, \ldots, T_d\}\) is the diffusion timestep, \(\alpha_t \in (0, 1)\) is a fixed noise schedule, and \(I\) is the identity matrix.
This process gradually transforms clean latent sequences into Gaussian noise, approaching a standard Gaussian \( \mathcal{N}(0, I) \) as \(\alpha_t\) decreases over time.

To reverse the noising process, we employ a transformer-based diffusion model \(F_\theta\) that predicts the clean latent sequence \(\hat{x}_0\) at each step:
\begin{equation}
\hat{x}_0 = F_\theta(x_t, y, t).
\end{equation}
While this corresponds to the \(x_0\)-prediction parameterization, it can also be interpreted in a score-based framework~\cite{song2020score, ddpm}, 
where the estimated score of the noisy latent distribution is computed as
\begin{equation}
\boldsymbol{\epsilon}_\theta(x_t, y, t) = \frac{x_t - \sqrt{\bar{\alpha}_t} \hat{x}_0}{\sqrt{1 - \bar{\alpha}_t}} 
\;\; \approx \;\; \nabla_{x_t} \log p_t(x_t \mid y).
\end{equation}
This interpretation naturally allows the integration of additional guidance signals during inference~\cite{song2020score}, as such signals can be added directly to the estimated score to steer the denoising trajectory. 
The diffusion model is trained to minimize the reconstruction error in the latent space:
\begin{equation}
\mathcal{L}_{\text{diff}} = 
\mathbb{E}_{x_0 \sim q(x_0),\ t \sim [1, T_d]}
\left[\|F_\theta(x_t, y, t) - x_0\|_2^2\right],
\end{equation}
where \( q(x_0) \) denotes the distribution of clean follower latent sequences in the training set.
The VQ-VAE modules are kept frozen during the training of the diffusion model.

\noindent
\textbf{Contact-guided Sampling.}
During inference, we further introduce contact guidance~\cite{yang2024smgdiff, song2020score} to enforce physical consistency.
Specifically, when the generated contact matrix indicates that two joints should be in contact, their spatial distance is encouraged to remain small, ensuring alignment between contact states and geometry. At each denoising step, a contact consistency loss $\mathcal{L}_{\text{c}}$ is computed as:
\begin{equation}
\mathcal{L}_{\text{c}} = 
\frac{\sum_{i,j} \| p^f_i - p^l_j \|_2^2 \cdot c_{ij}}
{\sum_{i,j} c_{ij} + \epsilon},
\end{equation}
where $p^f_i$ and $p^l_j$ are 3D joint positions of the follower and leader, $c_{ij}$ is the predicted contact mask, and $\epsilon$ is a small constant for numerical stability.

To integrate this constraint into the sampling process, we adopt a score-based guidance strategy~\cite{song2020score}, 
modifying the conditional score of the diffusion model at each step:
\begin{equation}
\nabla_{\mathbf{x}_t} \log p_t(\mathbf{x}_t \mid y) 
:= \nabla_{\mathbf{x}_t} \log p_t(\mathbf{x}_t \mid y) - \lambda_{\text{c}} \, \nabla_{\mathbf{x}_t} \mathcal{L}_{\text{c}},
\end{equation}
where $\lambda_{\text{c}}$ controls the strength of the guidance. 
In the DDIM~\cite{ddim} update, the modified score is used to compute the next latent step $\mathbf{x}_{t-1}$. This gradient correction guides the denoising trajectory toward motions that are consistent with the predicted contacts while respecting the conditional distribution of the leader motion and music features.

\section{Experiment}

\begin{table*}[t]
\caption{\textbf{Quantitative comparison with state-of-the-art methods on the DD100 dataset.} $\rightarrow$ indicates that a value closer to the Ground Truth is better. The best and the second-best results are highlighted in \textbf{bold} and \underline{underlined}, respectively.}
\centering
\resizebox{0.8\textwidth}{!}{ 
\begin{tabular}{l  c c c c c c c c c}
    \toprule
    &  \multicolumn{4}{c}{Solo Metrics} & \multicolumn{4}{c}{Interactive Metrics} & Rhythmic \\
    \cmidrule(r){2-5}  \cmidrule(r){6-9}
    Method & FID$_k$($\downarrow$)  & FID$_g$($\downarrow$)  & Div$_k$($\rightarrow$)  & Div$_g$($\rightarrow$)  
           & FID$_{cd}$($\downarrow$) & Div$_{cd}$($\rightarrow$) & CF(\%) & BED($\uparrow$)   & BAS($\uparrow$) \\
    \midrule
    Ground Truth  & 6.56 &  6.37 & 11.31 & 7.61 & 3.41 & 12.35 & 74.25 & 0.5308 & 0.1839 \\
    \midrule
    ReGenNet~\cite{regennet} & 9326.8 & 554.92 & 75.4 & 21.06 & 904.88 & 22.55 & 15.55 & \underline{0.3740} & \textbf{0.2086} \\
    InterFormer~\cite{interformer}  & 
    879.46 & 872.99 & 24.67 & 14.96 & 18.29 & \underline{10.88} & - & 0.1966 & 0.1848 \\
    Duolando~\cite{duolando} & 
    \underline{25.30} & \underline{33.52} & \textbf{10.92} & \underline{7.97} & \underline{9.97} & \textbf{14.02} & \underline{52.36} & 0.2858 & 0.2046 \\
    RCDiff & 
    \textbf{8.89} & \textbf{32.28} & \underline{9.53} & \textbf{7.35} & \textbf{8.01} & 10.42 & \textbf{61.58} & \textbf{0.4606} & \underline{0.2050} \\
    \bottomrule
\end{tabular}
}
\label{table:quantitative_eval}
\vspace{-1em}
\end{table*}

We evaluate our method through quantitative and qualitative comparisons with state-of-the-art methods, along with ablation studies to validate each component.
\begin{figure*}[t]
  \centering
  \includegraphics[width=\linewidth]{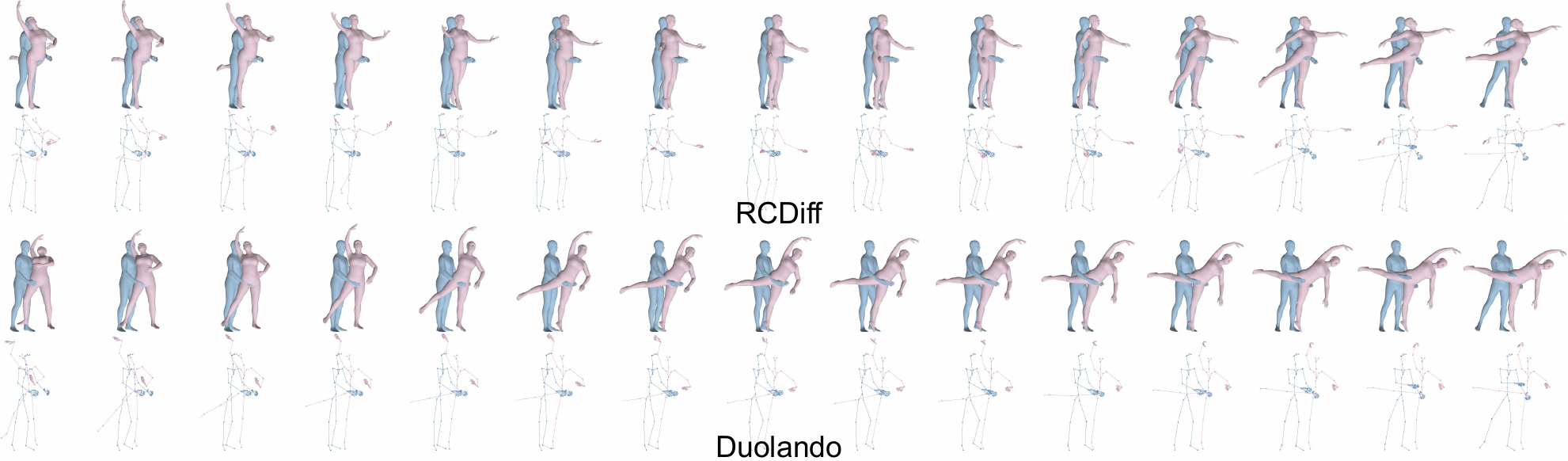} 
  \caption{\textbf{Qualitative comparisons of Duolando~\cite{duolando} and our method RCDiff.} The blue mesh represents the leader, while the pink mesh shows the generated follower, corresponding to the reactive motion produced by the models.}
  \label{fig:vis}
\end{figure*}

\subsection{Experimental Setup}

\noindent
\textbf{Datasets.} We evaluate our model on the newly proposed DD100 dataset~\cite{duolando}, a high-quality collection of duet dance interactions represented using the SMPL-X~\cite{smplx} body model. DD100 covers ten distinct dance styles, all characterized by strong physical interactions between dancers. The dataset features performances from five pairs of professional dancers, with each dance sequence accompanied by unique background music. In total, DD100 contains approximately 3.24 hours of recordings, which are randomly split into 80\% for training and 20\% for testing.

\noindent
\textbf{Metrics.} Following previous works~\cite{duolando, aist++, regennet}, we evaluate our model from three aspects: 
1) the quality of the generated follower motion,
2) the coordination between the follower and leader motions,
3) the alignment between the follower motion and the background music. 
For the first aspect, we compute the Frechet Inception Distance (FID) and diversity (Div) using geometric (denoted as ``g")~\cite{muller2005efficient} and kinetic (denoted as ``k")~\cite{onuma2008fmdistance} features. For the second aspect, we extract cross-distance features~\cite{duolando}, which encode the pairwise distances between ten selected joints of the leader and the follower at each frame, and compute \(\text{FID}_{cd}\) and \(\text{Div}_{cd}\) based on these features. Additionally, we compute contact frequency (CF) and Beat Echo Degree (BED)~\cite{duolando} to evaluate interaction strength and rhythmic consistency between the two dancers. For the third aspect, we use Beat-Align Score (BAS)~\cite{bailando} to assess the alignment between the generated motion and the background music.

\noindent
\textbf{Implementation Details.} 
For the PartFusion-VQ, we use the Adam optimizer~\cite{adam} with $\beta_1 = 0.5$ and $\beta_2 = 0.999$. 
The model is trained for 500 epochs with a batch size of 128 and 1000 iterations per epoch. 
The learning rate is initialized to $3 \times 10^{-5}$ and decayed by a factor of 0.1 at the 100th and 200th epochs. 
For the Contact VQ-VAE, we adopt the same optimizer and batch size as PartFusion-VQ. 
It is trained for 200 epochs, with the learning rate decayed by 0.1 at the 100th epoch. 
For the Trajectory VQ-VAE, since we follow the definition in~\cite{duolando}, we directly use the pre-trained model parameters provided in their work. 
For the diffusion model RCDiff, we use the AdamW optimizer~\cite{adamw} with a learning rate of $1.0 \times 10^{-4}$, a batch size of 16, and 150 epochs with 1000 iterations per epoch. 

\subsection{Comparison with the State-of-the-Arts}

We compare our model with existing reactive motion generation methods designed for two-person interaction: Duolando~\cite{duolando}, InterFormer~\cite{interformer}, and ReGenNet~\cite{regennet}.  
Duolando is a two-stage method for dance reaction generation. It first encodes motion and relative translations using VQ-VAEs, and then autoregressively generates the motion of the follower in the latent space with a GPT~\cite{gpt} model. For comparison, we use the quantitative results reported in the original paper, including the ground truth metrics they provide.
InterFormer and ReGenNet are both single-stage methods that predict motion in the space of 3D joint positions or rotations. InterFormer employs a transformer architecture, whereas ReGenNet is based on a diffusion model. We train and evaluate both methods on DD100 using their official implementations. The CF metric is omitted for InterFormer, as its output does not include joint rotations.

\begin{table*}[t]
\caption{\textbf{Contact Matrix Influence on Interaction Quality and Motion Detail Prediction.} $\rightarrow$ indicates that a value closer to the Ground Truth is better. The best and the second-best results are highlighted in \textbf{bold} and \underline{underlined}, respectively.}
\centering
\resizebox{0.8\textwidth}{!}{ 
\begin{tabular}{l  c c c c c c c c c}
    \toprule
    &  \multicolumn{4}{c}{Solo Metrics} & \multicolumn{4}{c}{Interactive Metrics} & Rhythmic \\
    \cmidrule(r){2-5}  \cmidrule(r){6-9}
    Method & FID$_k$($\downarrow$)  & FID$_g$($\downarrow$)  & Div$_k$($\rightarrow$)  & Div$_g$($\rightarrow$)  
           & FID$_{cd}$($\downarrow$) & Div$_{cd}$($\rightarrow$) & CF(\%) & BED($\uparrow$)   & BAS($\uparrow$) \\
    \midrule
    Ground Truth  &  6.56 & 6.37 & 11.31 & 7.61 & 3.41 & 12.35 & 74.25 & 0.5308 & 0.1839 \\
    \midrule
    w/o Contact  & 
     \underline{8.28} & 40.55 & 9.10 & 8.24 & 9.78 & 10.01 & 51.34 & \underline{0.4377} & 0.1958 \\
    w/o Contact Guidance & 
     \textbf{7.45} & \underline{32.87} & \underline{9.44} & \textbf{7.66} & \textbf{3.41} & \textbf{11.19} & \underline{60.23} & 0.4269 & \underline{0.2009} \\
    RCDiff & 
    8.89 & \textbf{32.28} & \textbf{9.53} & \underline{7.35} & \underline{8.01} & \underline{10.42} & \textbf{61.58} & \textbf{0.4606} & \textbf{0.2050} \\
    \bottomrule
\end{tabular}
}
\label{table:contact}
\end{table*}

\noindent
\textbf{Quantitative comparisons.}
\cref{table:quantitative_eval} shows that our method achieves the best overall performance.
For solo metrics, our model produces follower motion with much higher realism, reaching an FID$_k$ of 8.89 and an FID$_g$ of 32.28, improving over Duolando by 16.41 and 1.24, and exceeding InterFormer and ReGenNet by large margins. 
For diversity, our Div$_k$ is slightly lower than Duolando, while our Div$_g$ is closest to the ground truth.
For interaction quality, our method achieves an FID$_{cd}$ of 8.01, outperforming Duolando by 1.96. In contrast, ReGenNet exhibits extremely high interaction errors, with an FID$_{cd}$ of 904.88, far above our result. InterFormer also falls behind, with an FID$_k$ of 879.46, exceeding our score by 870.46.
Finally, for rhythmic alignment, our BED reaches 0.4606, improving over Duolando by 0.1748, and our BAS reaches 0.2050, which is higher than Duolando and ranks as the second-best result.

Notably, ReGenNet and InterFormer struggle with long sequences in DD100~\cite{duolando}. ReGenNet’s short segments must be concatenated, which harms continuity, and its absolute-position predictions over DD100’s large motion range cause gradual drift over time. InterFormer suffers from unexpected EOS events, also reducing output quality. In contrast, Duolando and our method operate in latent space, where VQ-VAE decoders naturally smooth sequences, producing more coherent motion.

\noindent
\textbf{Qualitative comparisons.}
Since Duolando achieves the closest quantitative performance to our method, we focus the qualitative comparison on it (Fig.~\cref{fig:vis}). The blue and pink meshes and skeletons represent the leader and generated follower, respectively, with meshes showing joint rotations and skeletons showing predicted joint positions. Overall, our method produces smooth, natural follower motions well coordinated with the leader, whereas Duolando often generates visually and physically implausible poses, such as abnormal wrist bending.

\subsection{Ablation Study}

\begin{figure}[t]
  \centering
  \includegraphics[width=.85\linewidth]{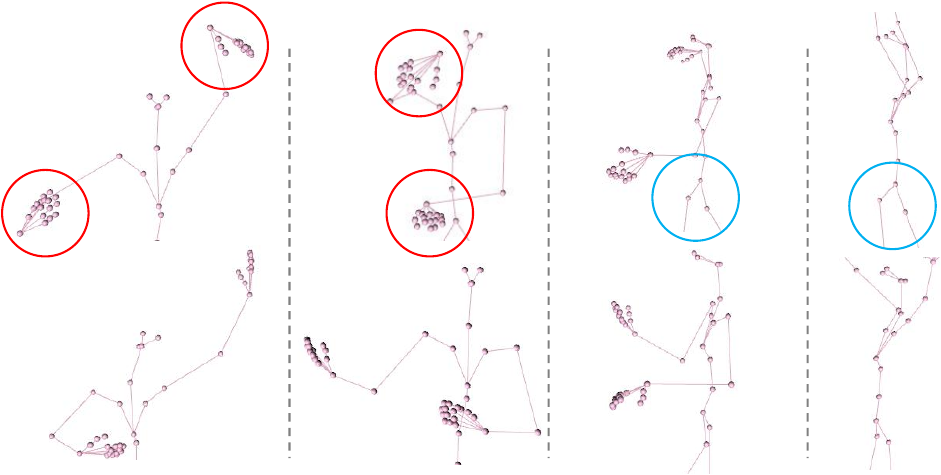} 
  \caption{\textbf{Ablation on PartFusion-VQ.} The first row shows the results obtained by modeling each body part separately with independent VQ-VAEs. The second row shows the results obtained by modeling the overall motion using PartFusion-VQ.}
  \label{fig:partvq_vis}
\end{figure}

\noindent
\textbf{Effectiveness of PartFusion-VQ.} 
We compare three variants: 1) a single VQ-VAE model, 2) separate VQ-VAEs for different body parts, and 3) PartFusion-VQ.
The comparison uses the metrics from ~\cite{ghosh2025duetgen}.
All models are trained within the same code framework for the same number of epochs.
As shown in \cref{table:vqvae}, PartFusion-VQ achieves the best performance. The relatively large errors highlight the intrinsic difficulty of joint reconstruction on DD100, due to its high diversity and complex motions, motivating the use of part-wise codebook modeling.
\begin{table}[t]
\caption{\textbf{Ablation on PartFusion-VQ.} The best and the second-best results are highlighted in \textbf{bold} and \underline{underlined}, respectively.}
\centering
\resizebox{.83\columnwidth}{!}{ 
\begin{tabular}{l c c c c}
    \toprule
    &  \multicolumn{2}{c}{MPJPE (mm) $\downarrow$} & \multicolumn{2}{c}{MPJVE (mm) $\downarrow$} \\
    \cmidrule(r){2-3}
    \cmidrule(r){4-5}
    Method & leader & follower & leader & follower \\
    \midrule
    single VQ-VAE & 149.28 & 164.44 & 18.87 & 21.23 \\
    separate VQ-VAEs &
    \underline{100.14} & \underline{100.67} & \textbf{12.65} & \underline{15.38} \\
    PartFusion-VQ & \textbf{80.32} & \textbf{86.78} & \underline{13.27} & \textbf{14.89} \\
    \bottomrule
\end{tabular}
}
\label{table:vqvae}
\end{table}
\cref{fig:partvq_vis} shows a comparison of generation results using separate VQ-VAEs and PartFusion-VQ.
In the two left columns, separate VQ-VAEs for the upper body, left hand, and right hand produce plausible local poses, but the connecting joints (\ie, wrists) exhibit unnatural bending. 
In the two right columns, separate VQ-VAEs for the upper and lower body cause the face and pelvis to point the same way, resulting in an anatomically implausible pose.
In contrast, PartFusion-VQ consistently produces plausible results thanks to the shared decoder.

\noindent
\textbf{Effectiveness of Contact Matrix.} 
To verify the effectiveness of the contact matrix, we perform an ablation study that includes removing it entirely from the diffusion model as well as using it only during training without applying contact-based guidance at test time.
As shown in~\cref{table:contact}, totally removing the contact matrix from the diffusion process degrades interaction quality, leading to lower \(\text{Div}_{cd}\), CF, and BED scores, while also increasing \(\text{FID}_{cd}\). Moreover, the absence of contact information also negatively impacts the prediction of fine-grained motion details for the follower, as reflected in an 8.27 increase in \(\text{FID}_{g}\). 
Applying contact guidance at test time further strengthens interaction consistency, helping the model enforce predicted contact patterns and leading to improved \text{BED}, \text{BAS}, and lower $\text{FID}_{g}$.

\subsection{User Study}

We conducted a user study to perceptually evaluate our method compared with Duolando~\cite{duolando}, the baseline closest to our method in quantitative metrics. 
25 participants of diverse ages, genders, and backgrounds, including varying familiarity with dance, were presented with 10 sets of anonymized animations.
In each set, two videos generated for the same leader sequence and music were shown side by side in randomized order. Participants rated each video on follower motion naturalness, follower-music alignment, and coordination with the leader using 5-point Likert scales (1 = worst, 5 = best).
Our method achieves higher scores than Duolando on all metrics: motion naturalness 4.26 vs 3.51, music alignment 4.24 vs 3.64, and coordination 4.31 vs 3.36, showing clear improvements in follower motion quality and interaction with the leader.

\section{Conclusion}
We propose a two-stage framework for reactive motion generation in duet dance, consisting of PartFusion-VQ and RCDiff. PartFusion-VQ encodes each body part with separate codebooks and a shared decoder, preventing unnatural poses while enabling fine-grained representation under limited data. RCDiff operates on this latent space, using diffusion to model complex motions, incorporating a contact matrix to improve leader-follower coordination, and applying contact guidance during inference to ensure plausible interactions. Experiments show that our method produces more realistic and interactive follower motions than others.
\noindent
\textbf{Acknowledgments:} This work was supported by ``Zhejiang Key Laboratory of Advanced Intelligent Warehousing and Logistics Equipment" (Grant No. 2024E10007)

\clearpage
{
    \small
    \bibliographystyle{ieeenat_fullname}
    \bibliography{main}
}


\end{document}